\pdfoutput=1

\documentclass[11pt]{article}

\usepackage[]{acl}

\usepackage{times}
\usepackage{latexsym}
\usepackage{graphicx}

\usepackage[T1]{fontenc}

\usepackage{booktabs}
\usepackage{multirow}
\usepackage{colortbl}

\usepackage[utf8]{inputenc}

\usepackage{microtype}

\usepackage{gb4e}

\usepackage{amsmath}
\usepackage{centernot}

\noautomath 

\title{What Makes Language Models Good-enough?}

\author{Daiki Asami \\
  University of Delaware \\
  \texttt{daiasami@udel.edu} \\\And
  Saku Sugawara \\
  National Institute of Informatics \\
  \texttt{saku@nii.ac.jp} \\}

\begin{document}

\maketitle

\begin{abstract}
Psycholinguistic research suggests that humans may build a representation of linguistic input that is `good-enough' for the task at hand.
This study examines what architectural features make language models learn human-like good-enough language processing.
We focus on the number of layers and self-attention heads in Transformers.
We create a good-enough language processing (GELP) evaluation dataset (7,680 examples), which is designed to test the effects of two plausibility types, eight construction types, and three degrees of memory cost on language processing.
To annotate GELP, we first conduct a crowdsourcing experiment whose design follows prior psycholinguistic studies.
Our model evaluation against the annotated GELP then reveals that the full model as well as models with fewer layers and/or self-attention heads exhibit a good-enough performance.
This result suggests that models with shallower depth and fewer heads can learn good-enough language processing.\footnote{Our dataset and codebase for dataset creation are available at \url{https://github.com/nii-cl/gelp}.}
\end{abstract}

\section{Introduction}
Language models exhibit impressive performance in various natural language understanding tasks \cite{devlin-etal-2019-bert, brown2020language, mahowald2023dissociating}, but one common concern is that they often rely on heuristics \cite{geirhos2020shortcut,du2023shortcut}.
For instance, BERT \citep{devlin-etal-2019-bert} makes predictions based on surface features, which leads to poor results in adversarial examples \citep{mccoy-etal-2019-right}.
Large language models such as GPT-2 \citep{radford2019language} also adopt fallible heuristics in in-context learning \citep{tang-etal-2023-large}.

\begin{figure}
    \centering
    \includegraphics[width=\linewidth]{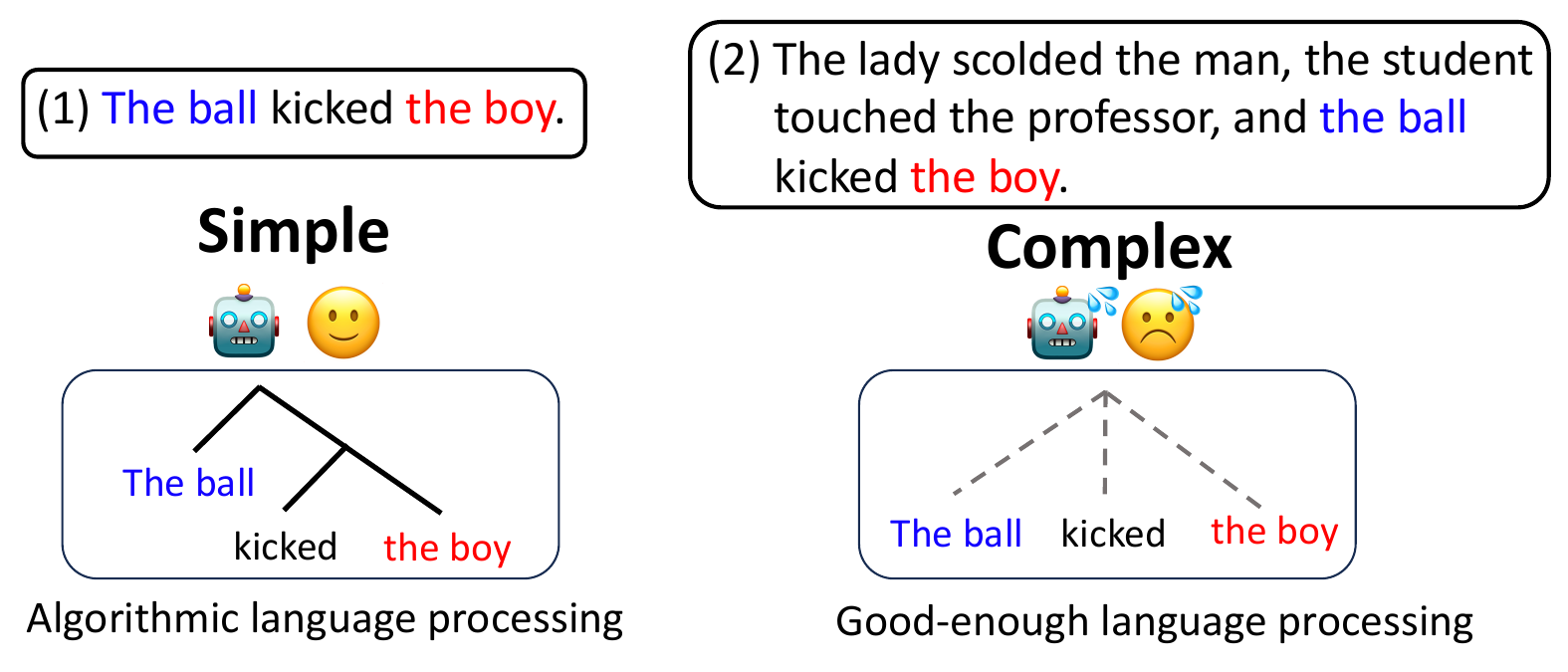}
    \caption{While algorithmic language processing involves a detailed syntactic analysis, good-enough language processing involves an incomplete one.
    We hypothesize that humans are more likely to adopt a good-enough strategy when processing a complex sentence than a simple one.
    A human-like good-enough model should exhibit a similar tendency.
    } \label{main_figure}
\end{figure}

\begin{table*}
\small
\centering 
\setlength{\tabcolsep}{17pt}
\begin{tabular}{ccc} \toprule
Construction & Implausible premise & Hypothesis (correct label)\\ \midrule
\multirow{2}{*}{(a) Transitive} & \multirow{2}{*}{\textcolor{blue}{The ball} kicked \textcolor{red}{the boy}.}
& \textcolor{red}{The boy}  kicked \textcolor{blue}{the ball}. (N)\\ & & \textcolor{blue}{The ball}  kicked \textcolor{red}{the boy}. (E)\\

\multirow{2}{*}{(b) Passive} & \multirow{2}{*}{\textcolor{red}{The boy} was kicked by \textcolor{blue}{the ball}.} & \textcolor{blue}{The ball} was kicked by \textcolor{red}{the boy}. (N)\\ & & \textcolor{red}{The boy} was  kicked by \textcolor{blue}{the ball}. (E)\\

\multirow{2}{*}{(c) DOC} & \multirow{2}{*}{The boy gave \textcolor{blue}{the apple} \textcolor{red}{the girl}.} & The boy gave \textcolor{red}{the girl} \textcolor{blue}{the apple}. (N)\\ & & The boy gave \textcolor{blue}{the apple} \textcolor{red}{the girl}. (E)\\

\multirow{2}{*}{(d) Dative} & \multirow{2}{*}{The boy gave \textcolor{red}{the girl} to \textcolor{blue}{the apple}.} & The boy gave \textcolor{blue}{the apple}  to \textcolor{red}{the girl}. (N)\\ & & The boy gave \textcolor{red}{the girl}  to \textcolor{blue}{the apple}. (E)\\

\multirow{2}{*}{(e) Ben. DOC} & \multirow{2}{*}{The cook made \textcolor{blue}{the bread} \textcolor{red}{the man}.} & The cook made \textcolor{red}{the man} \textcolor{blue}{the bread}. (N)\\ & & The cook made \textcolor{blue}{the bread} \textcolor{red}{the man}. (E)\\

\multirow{2}{*}{(f) Ben. \textit{for}} & \multirow{2}{*}{The cook made \textcolor{red}{the man} for \textcolor{blue}{the bread}.} & The cook made \textcolor{blue}{the bread}  for \textcolor{red}{the man}. (N)\\ & & The cook made \textcolor{red}{the man}  for \textcolor{blue}{the bread}? (E)\\

\multirow{2}{*}{(g) Exp. Subj.} & \multirow{2}{*}{\textcolor{blue}{The book}  liked \textcolor{red}{the girl}.} & \textcolor{blue}{The book}  liked \textcolor{red}{the girl}. (N)\\ & & \textcolor{red}{The girl}  liked \textcolor{blue}{the book}. (E)\\

\multirow{2}{*}{(h) Exp. Obj.} & \multirow{2}{*}{\textcolor{red}{The girl}  pleased \textcolor{blue}{the book}.} & \textcolor{red}{The girl} pleased \textcolor{blue}{the book}. (N)\\ & & \textcolor{blue}{The book}  pleased \textcolor{red}{the girl}. (E)
\\ \bottomrule
\end{tabular}
\caption{
    Eight constructions in GELP with example implausible premises and corresponding hypotheses.
    Abbreviations: Ben. = Benefactive; DOC = double object construction; Exp. = Experiencer; Obj. = Object; Subj. = Subject.
    }
\label{dataset examples}
\end{table*}

\begin{table*}
\small
\centering 
\setlength{\tabcolsep}{14pt}
\begin{tabular}{cc} \toprule
Memory load & Examples  \\ \midrule
Low (one proposition) & \textcolor{blue}{The ball}  kicked \textcolor{red}{the boy}. \\
Medium (two propositions) & The girl bought the cup and \textcolor{blue}{the ball}  kicked \textcolor{red}{the boy}. \\
High (three propositions) & The girl bought the cup, the singer broke the window, and \textcolor{blue}{the ball} kicked \textcolor{red}{the boy}.
\\ \bottomrule
\end{tabular}
\caption{
    Three degrees of memory load.
    The low, medium, and high memory load conditions include one, two, and three propositions, respectively.
    }
\label{memory_load_example}
\end{table*}

However, it is too hasty to view models' reliance on heuristics as a flaw.
According to a `good-enough' theory of human sentence processing \cite{ferreira2003misinterpretation,ferreira2007good,christianson2016language}, humans also adopt some types of heuristics.\footnote{Besides `good-enough', the psycholinguistic literature cited in the main text uses terms such as `heuristic', `shallow', and `underspecified'.
We use `good-enough' unless the difference is crucial.}
For instance, they may build an incomplete representation of linguistic input; as a result, they occasionally misinterpret an impossible description \cite[e.g., \textit{\textcolor{blue}{The ball} kicked \textcolor{red}{the boy}};][]{ferreira2003misinterpretation}.
The good-enough theory posits that such fallible language processing is good-enough for everyday communication.
In addition, it allows humans to efficiently process linguistic input by saving cognitive resources as suggested by the findings that they tend to rely on it when they face cognitive demands \citep[e.g., (1) vs. (2) in Figure~\ref{main_figure};][]{christianson2001thematic,christianson2006younger,christianson2010effects,patson2006individual}.

In light of this cognitive
background, we view language models' adaptation of heuristics as their potential for human-like linguistic performance \cite{linzen-2020-accelerate,hagendorff2023we}.
An open question here is what architectural features make them learn human-like good-enough language processing.
To study this question, we explore how the numbers of layers and self-attention heads affect models' performance.
Prior studies suggest that the model depth and attention heads contribute to syntactic generalizations \citep{mueller-linzen-2023-plant} and parallel a human working memory system \citep{ryu-lewis-2021-accounting,timkey-linzen-2023-language}, respectively.
Considering these findings, we hypothesize that (i) our aimed good-enough model requires a shallow depth because it does not have to engage in a detailed syntactic analysis and (ii) needs a small number of heads because a strong memory system is not necessary.

To test these hypotheses, we evaluate BERT's language processing capabilities through the lens of misinterpretation of sentences.
We create a good-enough language processing (GELP) dataset with 7,680 items.
GELP includes not only plausible but also implausible items to investigate humans' as well as models' misinterpretation.
Additionally, considering the prior psycholinguistic finding that some constructions are more likely to cause misinterpretation than others \cite{gibson2013rational}, it targets eight types of constructions (Table~\ref{dataset examples}).
Finally, to examine the effect of the memory demand, it operationalizes memory cost by including items with one, two, or three propositions (Table~\ref{memory_load_example}).

We first annotate the GELP dataset and probe humans' language processing by conducting a crowdsourcing experiment whose design follows previous psycholinguistic studies \citep{christianson2001thematic,christianson2006younger}.
Against the human data, we test which model with different architectures behaves in a human-like fashion.
We find that among 24 variants of BERT, those with fewer layers and heads perform in a good-enough way, similar to the full BERT-base model.
This result suggests that a deep architecture is not necessary for a model to learn good-enough language processing, which is consistent with our hypothesis (i).
In contrast, a closer look indicates that the role of attention heads in good-enough language processing does not confirm our hypothesis (ii).
This study modestly informs psycholinguistics by supporting the claim that humans' fallible language processing stems from a shallow syntactic analysis and suggesting that it has to do with the encoding phase of the working memory.

\section{Background and Motivation}
\subsection{Good-enough Theory in Psycholinguistics} \label{good_enough_theory}
A good-enough theory of language processing posits that humans may build representations that are good enough to achieve their communicative goal \citep{ferreira2003misinterpretation,ferreira2007good,christianson2016language}.
Such language processing is fallible in language use.
For instance, \citet{ferreira2003misinterpretation} find that humans misinterpret an implausible sentence (e.g., \textit{The professor bit the dog}) as its plausible version (e.g., \textit{The dog bit the professor}).
They reason that such misinterpretation results from canonical word order and plausibility heuristics.
Specifically, the plausible sentence is consistent with canonical patterns of English (i.e., a noun--verb--noun order generally represents an agent--action--patient relation) and world knowledge (i.e., dogs usually bite people, but not vice versa).
Another interpretation of humans' misinterpretation is that they may build a shallow representation of linguistic input \citep{sanford2002depth,sanford2006shallow}.
\citet{gibson2013rational} add to \citet{ferreira2003misinterpretation}'s finding by showing that the misinterpretation of implausible sentences depends on types of constructions.\footnote{\citet{gibson2013rational} do not adopt the good-enough theory but we introduce them because their findings are relevant to our study regardless of the theoretical framework that they adopt.}

Although good-enough language processing can cause errors, it enables humans to process language efficiently.
Previous studies suggest that such language processing is likely to take place under cognitively demanding tasks such as a priming paradigm involving both language comprehension and production \citep{christianson2010effects} and a reading experiment with structurally complex stimuli \citep{christianson2001thematic,ferreira2003misinterpretation}.
Additionally, other studies indicate that a limited working memory capacity can motivate humans to rely on a good-enough strategy \cite{christianson2006younger,patson2006individual}.
All of these findings are consistent with the view that humans adopt fallible but efficient good-enough language processing to reduce cognitive costs.

Crucially, humans can build a detailed representation if necessary.
For instance, misinterpretation does not easily occur in a situation that requires deep processing such as proofreading.
\citet{ferreira2003misinterpretation} emphasizes that humans' language processing has a good balance between robust, algorithmic language processing and non-robust, heuristic language processing. 

\subsection{Language Models' Reliance on Heuristics}
Heuristics also receive much attention in research on natural language understanding by language models \citep{geirhos2020shortcut,du2023shortcut}.
Language models are known to learn various types of heuristics based on training data during fine-tuning in natural language inference \citep{mccoy-etal-2019-right,gururangan-etal-2018-annotation}, question answering \citep{jia-liang-2017-adversarial,sugawara-etal-2018-makes,lai-etal-2021-machine}, and coreference inference \citep{zhao-etal-2018-gender} tasks.
Additionally, \citet{tang-etal-2023-large} find that non-fine-tuned large language models such as GPT-2 \citep{radford2019language} also adopt heuristics in in-context learning, suggesting that the recent models still suffer from the heuristics.

Despite this large body of research, it is an open question to what extent language models' reliance on heuristics resembles humans'.
This question is crucial because implementing human-like heuristics into language models can lead to a more efficient system in terms of resource and computational requirements \citep{hagendorff2023we}.

To tackle this question, we explore what architectural features contribute to the aimed model by focusing on the numbers of layers and self-attention heads in Transformers \citep{vaswani2017attention}.
We target these two architectural features because of their putative resemblance to human linguistic as well as non-linguistic cognitive systems.
\citet{mueller-linzen-2023-plant} show that model depth is important for models to learn syntactic generalizations.
Assuming the psycholinguistic claim that good-enough language processing does not involve a detailed syntactic analysis \citep{sanford2002depth,sanford2006shallow,ferreira2003misinterpretation,ferreira2007good,christianson2016language}, our first hypothesis (i) is that our aimed model does not require a deep architecture.
Regarding the self-attention heads, \citet{ryu-lewis-2021-accounting} and \citet{timkey-linzen-2023-language} suggest that a self-attention mechanism exhibits similarity to the retrieval phase of the working memory, but it remains open whether it also captures other phases (i.e., encoding and maintenance).
Given that adaptation of heuristics may result from demand on the working memory system \citep{christianson2006younger,patson2006individual}, our second hypothesis (ii) is that models with fewer heads behave in a human-like fashion, capturing humans' reliance on a good-enough strategy as a function of the memory demand.

\section{Dataset Creation}
To probe models' language processing, we create a natural language inference (NLI) dataset called GELP, targeting two plausibility types, eight types of constructions, and three degrees of memory load.
In an NLI task, models predict whether a sentence (\textit{premise}) entails, contradicts, or is neutral to another \citep[\textit{hypothesis};][]{condoravdi-etal-2003-entailment,bowman-etal-2015-large}.
For the human--model comparison discussed in Section~\ref{evaluation metric}, our labels include \textit{entailment} and \textit{non-entailment}, the latter of which covers both \textit{contradiction} and \textit{neutral}. 

\subsection{Low Memory Load Condition}
We first make items in the low memory load condition.
Considering the effect of construction types on misinterpretation discussed in Section~\ref{good_enough_theory} \cite{gibson2013rational}, GELP targets eight constructions (e.g., (a-h) in Table~\ref{dataset examples}).
We select 40 verbs from \citet{levin1993english} for (1) transitive/passive, (2) double object/dative, (3) benefactive double object/\textit{for}, (4) experiencer-subject, and (5) experiencer-object constructions each (a total of 200 verbs).\footnote{The complete list of verbs appears in Appendix~\ref{Verb_Lists}.}
These verbs take both animate and inanimate arguments within a single sentence, which allows us to make implausible sentences by swapping them.

With these verbs, we create 80 plausible premises for each construction (a total of 640 contexts) by giving GPT-3.5-turbo\footnote{https://platform.openai.com/docs/models} prompts.\footnote{We provide example prompts in Appendix~\ref{prompts}.}
We instruct it to make sentences with our selected verbs by using animate and inanimate nouns in positions of interest, which this paper indicates with red and blue for animate and inanimate nouns, respectively (e.g., \textit{\textcolor{red}{The boy} kicked \textcolor{blue}{the ball}}).
We manually check all generated sentences and correct any noticeable errors by hand (e.g., if an animate noun appears in an object position of \textit{kick}, we change it into an inanimate noun such as \textit{ball}).
Then, we swap the animate and inanimate nouns in each premise, creating 640 implausible premises (e.g., \textit{\textcolor{blue}{The ball} kicked \textcolor{red}{the boy}}).
Finally, we make two hypotheses with \textit{entailment} and \textit{non-entailment} labels for each premise.
As a result, the low memory condition has 2,560 pairs (\{8 constructions\}*\{80 premises\}*\{2 plausibility types\}*\{2 hypothesis types\}).

\subsection{Medium and High Memory Load Conditions}
Using the items in the low memory condition, we make pairs in the medium and high memory load conditions.
In doing so, we use templates, which allow us to create a large number of items systematically.
The medium memory load condition has templates for a premise and hypothesis such as (\ref{medium_memory_load_template}).

\begin{exe}
\ex \label{medium_memory_load_template} 
\begin{xlist}
\ex \label{medium_memory_load_context_template}
Target and the N1 V1 the N2.
\ex \label{medium_memory_load_question_template}
Entailed Hypothesis
\end{xlist}
\end{exe}

\noindent The premise consists of two propositions coordinated by one of the five connectives (\textit{and}, \textit{after}, \textit{when}, \textit{but}, and \textit{because}).
One proposition is a \textit{target} sentence (i.e., Target) that corresponds to a premise in the low memory condition, and the other is a template-generated \textit{filler} sentence (i.e., \textit{the N1 V1 the N2}).
For N1, V1, and V2 in the template, we use 201 transitive verbs and 515 animate nouns from \citet{fedorenko2020lack}.
Every selected noun is a plausible subject or object of every selected verb.
We ensure that these lexical items have no overlap with those used in target premises.
The target sentence either proceeds or follows the filler (e.g., \textit{Target and Filler} or \textit{Filler and Target}) to prevent one from developing a strategy to identify which proposition they should focus on while ignoring the other one.
A hypothesis can be either \textit{Entailed Hypothesis} or \textit{Non-entailed Hypothesis}, which correspond to hypotheses with \textit{entailment} and \textit{non-entailment} labels in the low memory condition, respectively.
Combining the five connectives, two target-filler premise orders, and two hypothesis types results in 80 templates.

A premise in the high memory load condition consists of three propositions coordinated by two of the five connectives used in the medium memory load condition.
We provide an example template for a premise in (\ref{high_memory_template}).

\begin{exe}
    \ex \label{high_memory_template}
    Target and the N1 V1 the N2 but the N3 V2 the N4
\end{exe}

\noindent
For a hypothesis, we use the same template as in the medium memory condition.
There are 20 permutation patterns of the two connectives out of the five connectives.
A target sentence appears in one of the three positions within the premise.
Combining the 20 connective patterns with the three proposition orders results in 60 templates.

Using the templates, we generate 2,560 pairs in the medium and high memory conditions each.
Consequently, GELP has a total of 7,680 items (2,560 pairs in three memory load conditions each).

\section{Human Experiment}

To annotate the GELP dataset and probe humans' language processing, we conduct a crowdsourcing experiment.
We explore to what extent plausibility types, construction types, and degrees of memory load lead to good-enough language processing.

\subsection{Methods} \label{human_methods}
We collect three responses per item in GELP on Amazon Mechanical Turk.\footnote{\url{https://www.mturk.com}}
We run our experiment on PCIbex.\footnote{\url{ https://farm.pcibex.net}}
Our results include data from 304 English native speakers.

Following previous psycholinguistic research \citep{christianson2001thematic,christianson2006younger,christianson2017reread}, our experiment uses a yes/no question-answering task instead of an NLI task, which our model evaluation uses.
We select this task because its procedure is simple to understand and natural to probe humans' language understanding.
In addition, the NLI task and yes/no question-answering task are highly interchangeable in the context of this study because to use GELP in the human experiment, all we have to do is to convert premises in GELP into polar questions (e.g., \textit{Did \textcolor{red}{the boy} kick \textcolor{blue}{the ball}?}), and to convert \textit{entailment} and \textit{non-entailment} labels into \textit{yes} and \textit{no} responses, respectively.
Because of this high interchangeability, we believe that the task difference does not hinder the human--model comparison.

\begin{figure}
    \centering    \includegraphics[width=7cm,height=5cm]{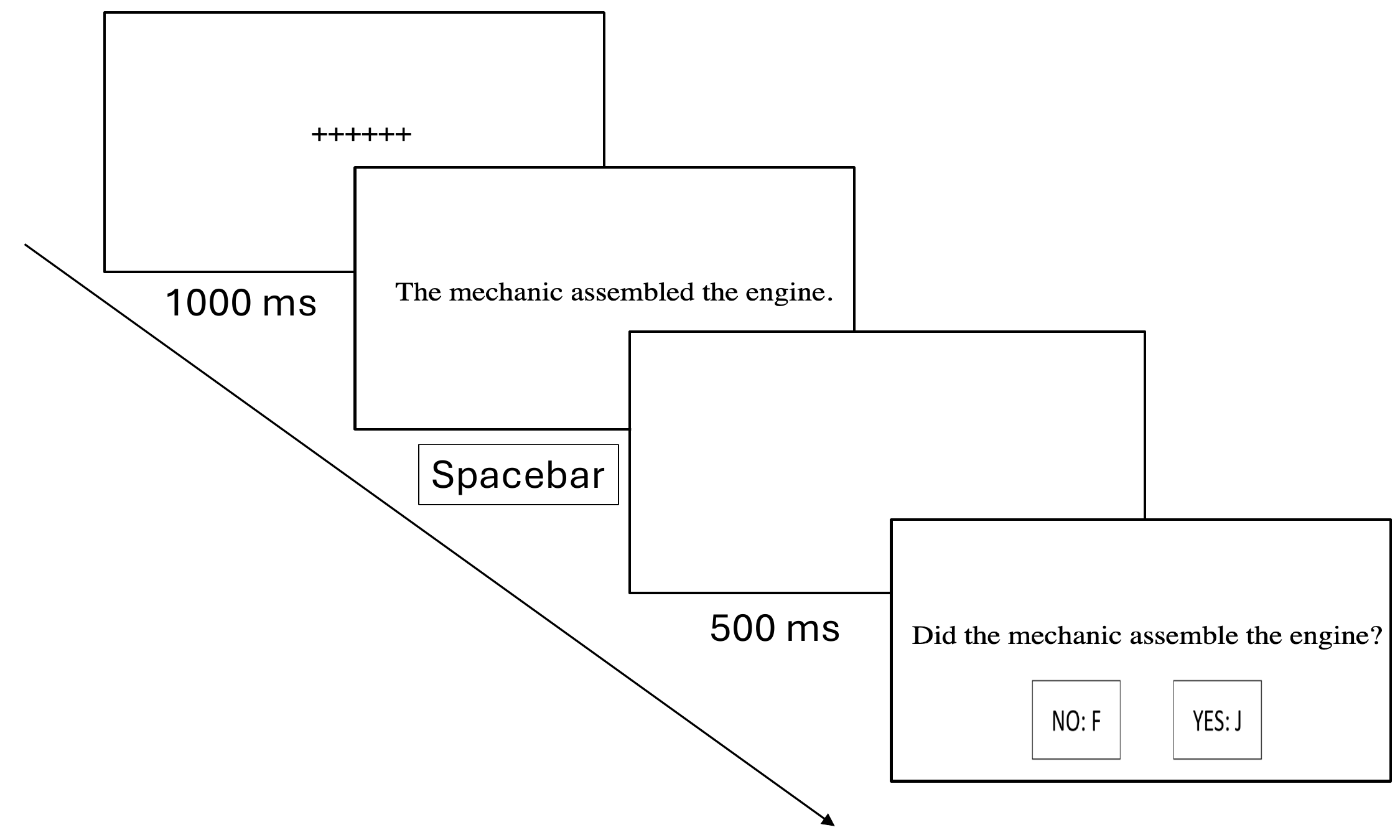}
    \caption{An example trial sequence in the human experiment.} \label{trial_sequance}
\end{figure}

Figure~\ref{trial_sequance} presents an example trial sequence in our experiment.
First, a 1,000 ms fixation occurs to solicit workers' attention.
Then, a context, which corresponds to a premise in GELP, appears in full.
A worker presses the spacebar to indicate that s/he has finished reading the sentence.
At this point, the sentence disappears.
After a 500 ms interval, a yes/no question appears, which the worker answers by pressing J or F for \textit{yes} or \textit{no}, respectively.
A worker repeats this procedure 96 times.\footnote{Appendix~\ref{Human_experiment_appendix} reports more details about the experimental methods.}

To calculate accuracy, we first assign each item a \textit{human answer} that represents the majority of the three responses and then determine whether it is equal to the pre-defined \textit{correct answer}.
The assignment of the human answer allows us to ensure that if an assigned human answer does not match a correct answer, the difference comes from misinterpretation rather than accident. 

Given previous psycholinguistic research reviewed in Section~\ref{good_enough_theory}, we expect that humans exhibit low accuracy due to good-enough sentence processing when they (i) process implausible descriptions, (ii) face a memory demand, and (iii) process certain implausible constructions relative to others.

\subsection{Results}
\begin{figure*}[t]
    \centering
    \includegraphics[trim=0 0 0 0.1cm,clip,width=\textwidth, height=9cm]{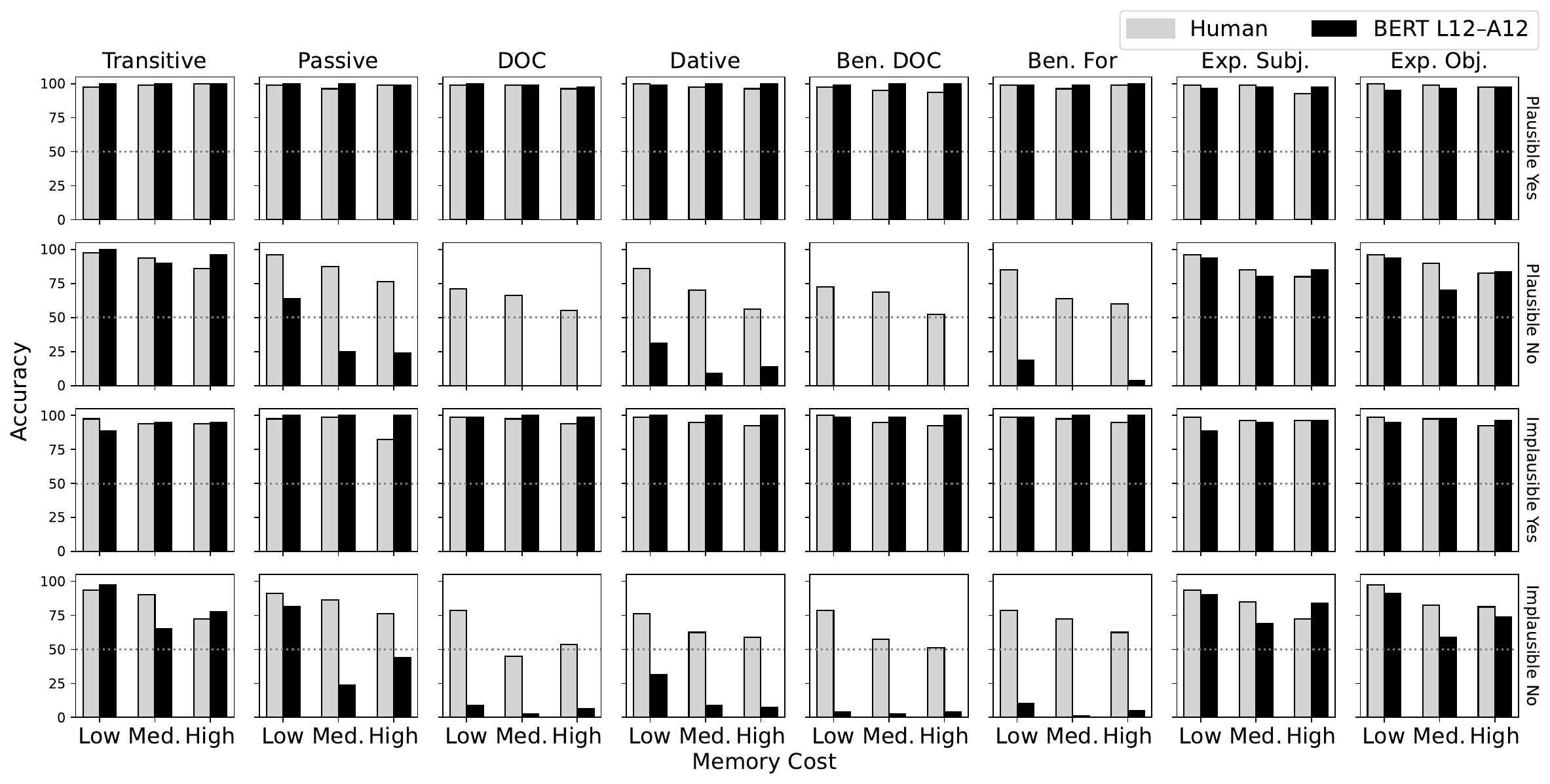}  
    \caption{Accuracy for humans and BERT L12--A12 on GELP.
    The dotted lines indicate chance-level performance (50\%).}
    \label{main_results}
\end{figure*}

The gray bars in Figure~\ref{main_results} represent accuracy for humans.
The average accuracy is 86.6\%.
This indicates that GELP is moderately challenging (cf. accuracy = 92 and 76\% on MNLI and Heuristic Analysis for NLI Systems (HANS), respectively, \citep{nangia-bowman-2019-human,mccoy-etal-2019-right}). 

Items with the correct \textit{yes} answer have higher accuracy than those with the correct \textit{no} answer (96.7 vs. 76.5\%).
We reason that humans may build a shallow syntactic representation of contexts; consequently, they tend to select \textit{yes} when the context and question exhibit word overlap.

We saw little difference between the items with plausible contexts and those with implausible ones (87.9 vs 85.4\%), contrary to our expectation (i).
This indicates that humans do not adopt canonical form--meaning mapping or plausibility heuristics presumably because they adopt only the shallow syntactic analysis throughout our experiment.

The accuracy decreases as the memory load increases (92.8, 86.2, and 80.9\% for low, medium, and high memory load conditions, respectively), confirming our expectation (ii).
This result suggests that shallow language processing comes from a task-related memory demand.
That is, humans adopt a shallow processing strategy to save cognitive resources so that they can memorize multiple propositions for subsequent question answering.

A closer look reveals that the accuracy is at the ceiling on the items with the correct \textit{yes} answer (min. = 94.5\%) but decreases on those with the correct \textit{no} answer.
The decrease depends on the memory load (86.9, 75.4, and 67.3\% for low, medium, and high loads, respectively).
This result should not be attributable to fallible memory.
This is because the accuracy on the items with the correct \textit{yes} answer is high regardless of the memory cost, indicating that humans can encode and store linguistic input in memory.
A more viable explanation of the result would be that they do not engage in a detailed syntactic analysis in the first place under a memory-demanding situation; as a result, they encode words but not a detailed syntactic representation in working memory.

Finally, the accuracy on each construction varies but the variation is not specific to implausible sentences, contrary to our expectation (iii).
Specifically, the accuracy on target sentences with two nouns is higher than that on those with three nouns (91.9 vs. 81.6\% on average for transitive, passive, experiencer subject, and experiencer object constructions, on the one hand, and double object, dative, benefactive double object, and benefactive \textit{for} constructions, on the other) regardless of plausibility (93.5 vs. 82.3\% and 90.2 vs. 80.5\% for plausible and implausible contexts of each group, respectively).
Unlike \citet{gibson2013rational}, we use items with more than one proposition and present the context and question separately.
We reason that the memory demand due to this task design facilitates a shallow syntactic analysis, which overrides another type of language processing observed in \citet{gibson2013rational}.
The similarity in the accuracy pattern between the plausible and implausible conditions is not surprising if humans adopt the shallow syntactic analysis rather than other strategies throughout the experiment.

In summary, we confirm our expectation (ii) that the accuracy drops as the memory cost increases, suggesting that humans adopt good-enough sentence processing in a memory-demanding situation.
However, we do not confirm the other two expectations (i) that implausible items have lower accuracy than plausible ones and (iii) that this difference depends on constructions.

\section{Model Evaluation}
Against the annotated GELP, we evaluate models with different architectural features.
Assuming that a good-enough model neither engages in a detailed syntactic analysis nor needs a strong working memory, we hypothesize that our aimed model has (i) a shallow architecture and (ii) a small number of self-attention heads relative to the full model. 

\subsection{Models}
We use Huggingface's \citep{wolf-etal-2020-transformers} 24 BERT miniatures \citep{turc2019}, which cross six numbers of layers (L $\in$ $\{2,4,6,8,10,12\}$) with four numbers of self-attention heads (A $\in$ $\{2,4,8,12\}$).
\citet{turc2019} set four hidden embedding sizes (H $\in$ $\{128, 256, 512, 768\}$) for each number of heads.
We hereafter denote models as BERT L$n$--A$n$ (e.g., BERT L12--A12).

Our evaluation uses an NLI task instead of the question-answering task because more training data are available for the former than the latter.
As noted in Section~\ref{human_methods}, the task difference should not hinder us from evaluating models' good-enough language processing.
We finetune the 24 BERT models on two standard NLI training datasets---SNLI \citep{bowman-etal-2015-large} and MNLI \citep{williams-etal-2018-broad}---and a training split of HANS \citep{mccoy-etal-2019-right}.
The inclusion of HANS intends to prevent the models from adopting structural heuristics all the time.
Without HANS, we find that the models perform at chance on GELP by predicting \textit{entailment} most of the time due to lexical overlap between a premise and hypothesis (e.g., 53\% accuracy for BERT-base fine-tuned without HANS).

\begin{table*}
  \centering
    \begin{tabular}{cccccc}
  \toprule
  L/A & 2 & 4 & 8 & 12 & Avg. \\
  \midrule
  2 & \cellcolor{gray!15}59.1 (0.6)& \cellcolor{gray!15}58.6 (0.6) & \cellcolor{gray!15}59.0 (0.6)& \cellcolor{gray!15}57.9 (0.6)& \cellcolor{gray!15}58.7 (0.6)\\
  4 & \cellcolor{gray!15}57.8 (0.6)& \cellcolor{gray!15}55.7 (0.6)& \cellcolor{gray!25}60.0 (0.6)& \cellcolor{gray}70.4 (0.5)& \cellcolor{gray!15}59. (0.6)\\
  6 & 54.3 (0.6)& 52.6 (0.6)& \cellcolor{lightgray}65.3 (0.5)& \cellcolor{lightgray}67.8 (0.5)& \cellcolor{gray!25}60.0  (0.6)\\
  8 & 54.3 (0.6)& \cellcolor{gray!15}58.9 (0.6)& \cellcolor{gray}70.0 (0.5)& \cellcolor{gray}71.4 (0.5)& \cellcolor{gray!25}63.7  (0.6)\\
  10 & 54.0 (0.6)& \cellcolor{lightgray}65.6 (0.5) & \cellcolor{gray}71.0 (0.5)& \cellcolor{gray}71.0 (0.5)& \cellcolor{lightgray}65.3 (0.5)\\
  12 & 49.9 (0.6)& \cellcolor{gray!25}61.4 (0.6)& \cellcolor{gray}73.2 (0.6)& \cellcolor{gray}74.3 (0.5) & \cellcolor{gray!25}64.7 (0.6)\\
  \hline
  Avg. & 54.9 (0.6)& \cellcolor{gray!15}58.8 (0.6)& \cellcolor{lightgray}66.4 (0.6) & \cellcolor{lightgray}68.8 (0.5)& \\
  \bottomrule
\end{tabular}
  \caption{Human--model matching score for 24 BERT models with different numbers of layers (L) and self-attention heads (A). Standard errors are in the parentheses.\label{model_score}}

\end{table*}

\subsection{Evaluation Metric} \label{evaluation metric}
Our evaluation metric is what we call a \textit{human--model matching score}.
This is calculated based on how many predicted labels match human responses.
\textit{Entailment} and \textit{non-entailment} labels in the model evaluation correspond to \textit{yes} and \textit{no} responses in the human experiment, respectively.
This compatibility between NLI labels and question-answering responses allows us to directly compare human and model results.
Importantly, the human--model matching score is not the same as accuracy.
For instance, the human--model matching score is 0.0 while the model accuracy is 1 if the model correctly labels the item while humans respond incorrectly.

\subsection{Results}
Table~\ref{model_score} presents the calculated human--model matching scores for each model.
Although not perfect, the best-performed model is BERT L12--A12 (matching score = 74.3\%), which corresponds to BERT-base.
Six models with fewer layers and/or heads (BERTs L\{4, 6, 8, 12\}--A\{8, 12\} excluding BERTs L4--A8, L6--A8 and L6--A12) show 70\% or higher scores (range: 70.0--73.2\%).
The comparable performance among these models suggests that good-enough language models do not require large architectures, confirming our overall hypothesis.

On average, the matching score improves from 54.9 to 68.7\% and from 58.7 to 64.7\% as layers and heads increase, respectively.
Importantly, the difference among Ls = 8, 10, 12 is small (63.7, 65.3, 64.7\%, respectively).
This suggests that a deep architecture is not necessary for a good-enough model, which is consistent with our expectation (i).
To explore this consideration, we finetune BERT L24--H16, which corresponds to BERT-large, in the same way as the 24 BERT models.
We find that the matching score and accuracy for this model are 76.7 and 73.4\%, respectively.
The 2.3 vs. 4.2\% improvement from BERT-base to BERT-large in these matrices suggests that a deep architecture contributes to the accuracy but less so to the human-like good-enough performance.
Although a more in-depth analysis is necessary, this finding seems to be consistent with our interpretation of the model depth.

To analyze the effect of the number of attention heads, we calculate the model accuracy on GELP based on the memory cost (Table~\ref{accuracy_by_memory_cost}).
We only report the results for models with 8 or 12 heads because most models with fewer heads perform at around the chance level.\footnote{Appendix~\ref{accuracy_by_memory_cost_appendix} presents the full results.}
The result shows that models exhibit decreasing accuracy as the memory cost increases.
However, this accuracy pattern does not depend on the number of attention heads.
For instance, the accuracy for the full BERT L12--A12 model was 74.1, 68.3, and 65.1\% on low, medium, and high memory load conditions, respectively.
This suggests that the number of attention heads does not necessarily contribute to models' good-enough performance, which is inconsistent with our hypothesis (ii).
To further explore this conjecture, we fine-tune BERT-large, which corresponds to BERT L24--A16, in the same way as 24 BERT models.
The fine-tuned BERT-large shows a similar pattern as the BERT L12--A12: the accuracy decreases as the memory load increases (77.9, 77.1, and 75.0\% on low, medium, and high memory load conditions, respectively) but the decrease is more moderate than that observed in BERT-base.
Since BERT-large differs from BERT-base in terms of hyperparameters other than the number of heads such as the model depth, we leave open what leads to this moderate decrease.

\begin{table*}
\centering 
    \begin{tabular}{cccccccc}
\toprule
{L/A} & \multicolumn{3}{c}{8} & \multicolumn{3}{c}{12} \\
\cmidrule(lr){2-4} \cmidrule(lr){5-7}
{Memory load} & {Low} & {Medium} & {High} & {Low} & {Medium} & {High} \\
\midrule
2  & 50.9 (2.8)& 50.6 (2.8)& 50.2 (2.8)& 49.1 (2.8)& 50.5 (2.8)& 49.8 (2.8)\\
4  & 54.7 (2.8)& 51.9 (2.8)& 52.9 (2.8)& \cellcolor{lightgray}69.4 (2.6)& \cellcolor{gray!25}64.8 (2.7)& \cellcolor{gray!25}63.6 (2.7)\\
6  & \cellcolor{gray!15}59.0 (2.8)& \cellcolor{gray!15}58.0 (2.8)& \cellcolor{gray!15}56.6 (2.8)& \cellcolor{gray!25}62.7 (2.7)& \cellcolor{gray!25}61.8 (2.7)& \cellcolor{gray!25}61.4 (2.7)\\
8  & \cellcolor{lightgray}66.1 (2.7)& \cellcolor{gray!25}64.8 (2.7)& \cellcolor{gray!25}61.3 (2.7)& \cellcolor{lightgray}69.2 (2.6)& \cellcolor{lightgray}66.3 (2.7)& \cellcolor{gray!25}64.8 (2.7)\\
10 & \cellcolor{lightgray}66.3 (2.6)& \cellcolor{lightgray}65.6 (2.7)& \cellcolor{gray!25}64.6 (2.7)& \cellcolor{lightgray}68.5 (2.6)& \cellcolor{gray!25}64.6 (2.7)& \cellcolor{gray!25}62.9 (2.7)\\
12 & \cellcolor{lightgray}69.3 (2.6) & \cellcolor{lightgray}68.1 (2.6) & \cellcolor{lightgray}67.7 (2.6) & \cellcolor{gray}74.1 (2.5)& \cellcolor{lightgray}68.3 (2.6)& \cellcolor{lightgray}65.1 (2.7)\\
\midrule
Avg. & \cellcolor{gray!25}61.1 (2.7)& \cellcolor{gray!15}59.8 (2.7)& \cellcolor{gray!15}58.9 (2.7)& \cellcolor{lightgray}65.5 (2.6)& \cellcolor{gray!25}62.7 (2.7)& \cellcolor{gray!25}61.3 (2.7)\\
\bottomrule
    \end{tabular}
    \caption{Accuracy for BERT with 8 or 12 heads on GELP based on the three degrees of memory cost. Standard errors are in the parentheses.} \label{accuracy_by_memory_cost}
\end{table*}

BERT L12--A12 shows the best matching score, 74.3\%, and we find that its accuracy based on the pre-defined correct label is 69.2\%, which is below human accuracy 86.6\%.
To disentangle what makes the model similar or different from the humans, Figure~\ref{main_results} presents the accuracy for the BERT L12--A12 on each condition in the black bars.
We find that it performs at the ceiling on items with correct \textit{entailment} (= \textit{yes}) labels in a similar way as humans.
However, its performance on items with correct \textit{non-entailment} (= \textit{no}) depends on construction types.
Specifically, BERT L12--A12 performs well on the constructions with two-place predicates (i.e., transitive, experiencer subject, and experiencer object constructions; 83.3\%) but at around chance on passives (43.5\%) and poorly on constructions with three-place predicates (double object, dative, benefactive \textit{for}, and benefactive double object constructions; 7.0\%).
The model's performance in the latter two does not resemble humans' although they are also not perfect (85.6 and 66.0\% for passives and three-place predicate constructions, respectively).

The models' poor performance on passives and three-place predicates has a broad implication.
Recent studies suggest that language models can learn to be sensitive to word order \cite{papadimitriou-etal-2022-classifying,kauf2023event}.
However, our results indicate that models' word order sensitivity can be specific to active sentences with two-place predicates.
To explore whether the observed poor performance comes from the model's internal architecture or lack of relevant examples in training data, we retrain BERT L12--H12 on data augmented with 800 examples similar to passives or ditransitive constructions in GELP.
The retrained model achieves the 84.3\% human--model matching score.
Although we cannot draw a strong conclusion because the training data that we use resembles the items in GELP, this result suggests that use of appropriate training data leads models to learn robust syntactic generalizations \citep{mccoy-etal-2019-right}.

In summary, the full model as well as models with fewer layers and/or heads show good-enough performance.
The smaller number of layers does not considerably impair the models' good-enough language processing, confirming our first hypothesis (i) that 
a shallow architecture leads to a shallow representation of linguistic input.
In contrast, the contribution of the number of heads is unclear, which does not confirm our second hypothesis (ii) that fewer heads parallel humans' limited working memory system.

\section{Discussion}
\paragraph{Does a shallow architecture lead to good-enough language processing?} Our first hypothesis is that a good-enough model requires a shallow architecture because it does not have to make a detailed syntactic representation of linguistic input.
We confirm this hypothesis by finding that the shallow models exhibit a human-like good-enough performance in a way similar to their deeper version.
Specifically, increasing the number of layers from 8 to 12 does not improve the models' human-like performance considerably.

Our results shed light on whether language models can learn the dissociation between formal linguistic competence---knowledge of grammar---and functional linguistic competence---the ability to use language in real-world situations \citep{mahowald2023dissociating}.
The advent of seemingly well-behaved neural language models leads to an intensive investigation of their formal linguistic competence \citep{marvin-linzen-2018-targeted,futrell-etal-2019-neural,warstadt-etal-2019-neural,warstadt-etal-2019-investigating,warstadt-etal-2020-blimp-benchmark}.
\citet{mahowald2023dissociating} conclude that language models show promising results in learning abstract linguistic rules and patterns, but it remains to be seen whether they can learn functional linguistic competence.
Our results point to the possibility that deep as well as shallow models can learn human-like language processing that is good-enough for simple language use.

\paragraph{Do fewer heads lead to good-enough language processing?} \label{heads_discussion} Our second hypothesis is that the aimed model requires fewer attention heads because it does not need a strong working memory system.
We find that among our model set, the model with the largest number of heads (H = 12) shows decreasing accuracy as the memory load increases, in a similar way as humans.
This result does not confirm our hypothesis.

We can explain this result if the parallelism between the self-attention mechanism in Transformers and the human working memory system is specific to the retrieval phase.
The human working memory system involves three phases, encoding, storage, and retrieval, at a coarse level of granularity \citep{baddeley1986working,baddeley2000episodic}. 
The models as well as humans achieve accuracy at the ceiling on items with correct \textit{entailment}/\textit{yes} labels, suggesting that they can store words that appear in contexts/premises.
Thus, the observed accuracy pattern might not have to do with the storage phase of the working memory.
We then conjecture that good-enough language processing has to do with the encoding phase of the working memory: the detailed syntactic analysis does not take place in the first place due to a high memory load during this phase, and as a result, the working memory does not store the detailed syntactic representation.
We leave open the exact mechanism of this process and its connection to the self-attention architecture in Transformers for future research.

\paragraph{How does this study inform psycholinguistics?}
Although it is hard to make a direct comparison between humans and language models, it is worth considering if our findings can inform psycholinguistics in a meaningful way.
Prior psycholinguistic studies postulate multiple possible sources of humans' good-enough performance.
Some representative examples are semantic or structural heuristics \citep{ferreira2003misinterpretation}, a shallow representation of linguistic input \citep{sanford2002depth,sanford2006shallow}, and working memory burden \citep{christianson2006younger,patson2006individual}.
The results from our model evaluation against human data are at least consistent with the view that humans adopt a non-detailed syntactic analysis.
As we stipulated in the preceding paragraph, they also point to the possibility that it is the encoding but not storage or retrieval phase of the working memory system that is related to a source of humans' shallow sentence processing.
Combining these two considerations suggests that the working memory demand during the encoding phrase leads to the shallow syntactic analysis of sentences \citep[cf.][]{christianson2006younger}.

\section{Conclusion}
From a cognitive perspective, we take language models' heuristic performance as their potential to learn human-like good-enough performance in language processing.
This study creates a good-enough language processing evaluation dataset, GELP.
We explore what architectural features contribute to models' good-enough language processing, focusing on the numbers of layers and self-attention heads.
The model evaluation reveals that models with fewer layers and/or heads exhibit a good-enough performance in a similar way as a full model.
This result leads us to conclude that the shallow architecture makes the models engage in an underspecified syntactic analysis.
In contrast, the role of the self-attention mechanism is unclear.
We leave open its exact role in the model's good-enough language processing.
We hope that this study will foster more psycholinguistically oriented research on language models' good-enough performance.

\section*{Limitations}
This study has three limitations.
First, our model coverage is limited.
We evaluate only BERT with different architectural features.
Therefore, it remains to be seen whether our findings and hypotheses are generalizable to other Transformer-based models such as GPT-2.

Second, the use of templates in the dataset creation may result in unnatural sentences in an unintended way such that it leads us to fail to measure what we intend to measure (e.g., the effect of the construction types).

Finally, this study does not test whether models can perform in a human-like way in terms of both grammatical knowledge and good-enough language processing.
According to \citet{ferreira2003misinterpretation}, humans have a good balance between robust, algorithmic language processing and non-robust, heuristic language processing so that they do not always misinterpret linguistic input.
It is necessary to test whether models can learn the human-like balance.

\section*{Ethics Statement}
When we use Amazon Mechanical Turk, we make sure that our payment and rejection policies are reasonable and comparable to in-person employment.
The task that we use in our crowdsourcing experiment is a simple yes/no question-answering task; hence, it should cause no harm to workers.
This work passes review from the oversight of the internal review boards of the authors' institutes.

\section*{Acknowledgements}
We wish to thank the anonymous reviewers for their constructive feedback.
This work was supported by JST PRESTO Grant Number JPMJPR20C4 and JSPS KAKENHI Grant Number 22K17954.

\bibliography{anthology,custom}
\bibliographystyle{acl_natbib}

\appendix

\section{Verb Lists} \label{Verb_Lists}
We use the following verbs for creating our dataset.

\paragraph{Verbs for transitives/passives}
\textit{assemble, bend, bite, blend, carve, chop, clean, collect, cut, describe, design, destroy, draw, flatten, fold, hack, hammer, hit, kick, knock, make, memorize, pound, produce, protect, punch, push, read, saw, save, shatter, skip, slash, slice, smash, squash, suggest, touch, use, waste}

\paragraph{Verbs for DOCs/datives}
\textit{allocate, assign, award, bring, email, extend, fax, feed, forward, give, grant, hand, haul, issue, lend, lease, leave, loan, mail, offer, owe, pass, pay, post, promise, refund, relay, repay, sell, send, serve, ship, show, slip, smuggle, take, teach, tell, trade, write
}

\paragraph{Verbs for benefactive DOCs/benefactive \textit{for} constructions}
\textit{arrange, assemble, bake, book, boil, build, buy, carve, cash, catch, charter, clean, compile, cook, cut, design, develop, earn, find, fix, get, grill, grow, keep, knit, make, order, paint, pick, prepare, rent, reserve, roll, save, secure, set, shape, steal, wash, write}

\paragraph{Verbs for experiencer subject constructions}
\textit{abhor, admire, adore, appreciate, cherish, covet, crave, deplore, desire, despise, disdain, dislike, distrust, dread, enjoy, envy, exalt, execrate, favor, fear, hate, lament, like, loathe, love, miss, mourn, need, pity, regret, relish, resent, savor, tolerate, treasure, trust, value, venerate, want, worship}

\paragraph{Verbs for experiencer object constructions}
\textit{agonize, amaze, amuse, anger, annoy, arouse, astonish, bore, bother, calm, captivate, comfort, confuse, convince, depress, devastate, disappoint, discourage, disgust, disturb, displease, embarrass, encourage, enlighten, excite, frighten, frustrate, impress, irritate, please, puzzle, sadden, satisfy, shock, surprise, terrify, threaten, thrill, upset, worry}

\section{Example Prompts} \label{prompts}
\paragraph{A template for a prompt:}
Can you make [CONSTRUCTION NAME] with the following verbs?
\\
\\
\noindent
Please...\\
1. Use an inanimate entity in [POSITION OF INTEREST].
\\
2. Use an animate entity in [POSITION OF INTEREST].\\
3. Use past tense for the verb.\\
4. Use no pronouns.\\
5. Use no adjectives.\\
\\
\noindent
[LIST OF VERBS]
\paragraph{An example prompt for the creation of transitive sentences:}
Can you make transitive constructions with the following verbs?
\\
\\
\noindent
Please...\\
1. Use an inanimate entity in the subject.\\
2. Use an animate entity in the object.\\
3. Past tense for the verb.\\
4. Use no pronouns.\\
5. Use no adjectives.\\
\\
agonize, amaze, amuse, anger, annoy, arouse, astonish, bore, bother, calm, captivate, comfort, confuse, convince, depress, devastate, disappoint, discourage, disgust, disturb, displease, embarrass, encourage, enlighten, excite, frighten, frustrate, impress, irritate, please, puzzle, sadden, satisfy, surprise, shock, terrify, threaten, thrill, upset, worry

\section{Details on Human Experiment} \label{Human_experiment_appendix}
\subsection{Participants}
Using Amazon Mechanical Turk, we recruit workers with the requirements of having an approval rating of 99.0\% or higher, having at least 5,000 approved tasks, and being located in the US, the UK, or Canada.
We calculate the reward as \$12 per hour.
We first recruit 1,200 workers for the qualification task.
The qualification task has 20 context--question pairs, which resemble items in GELP.
We take more than 70\% accuracy in the qualification task as a threshold for the invitation to the actual experiment.
Based on this exclusion criterion, we invite 487 workers to participate in the actual experiment.
Among them, 304 workers take part.

Our experiment collects no personal information.
By accepting PCIbex's participation agreement, workers consent to the collection and use of non-personal data for research purposes.

\subsection{List Design}
In addition to critical 7,680 items in GELP, our experiment includes 7,680 distractor pairs to ensure a balanced and unbiased assessment of humans' sentence processing.
They serve to mitigate potential response strategies (e.g., paying attention to only a target context) by masking the critical items.
Contexts in the distractors consist of two (N = 2,560) or three (N = 5,120) propositions, where one proposition corresponds to a target context; questions ask about a thematic relation in filler contexts.
Half of them have a correct \textit{yes} response; the other half have a correct \textit{no} response.

\begin{figure*}
    \centering
\includegraphics[width=6cm,height=20cm]{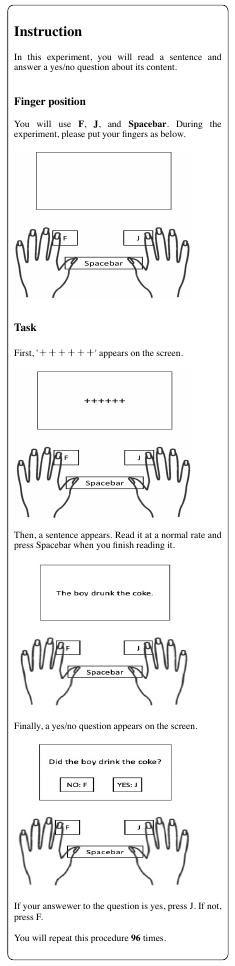}
    \caption{Instruction for the experiment.} \label{instruction}
\end{figure*}

We divide the 15,360 context--question pairs (7,680 targets + 7,680 distractors) into 160 lists, each of which has 96 items.
The number of items in each list roughly follows previous psycholinguistic research \citep{christianson2001thematic,christianson2006younger}.
We make sure that no worker sees the same list more than one time.

\subsection{Instruction}
We show workers an instruction in Figure~\ref{instruction} before the experiment to familiarize them with the experimental procedure.

\begin{table*}
\small
\centering 
    \begin{tabular}{cccccccccccccc}
\toprule
{L/A} & \multicolumn{3}{c}{2} & \multicolumn{3}{c}{4} & \multicolumn{3}{c}{8} & \multicolumn{3}{c}{12} \\
\cmidrule(lr){2-4} \cmidrule(lr){5-7} \cmidrule(lr){8-10} \cmidrule(lr){11-13}
{Memory load} & {Low} & {Med.} & {High} & {Low} & {Med.} & {High} & {Low} & {Med.} & {High} & {Low} & {Med.} & {High} \\
\midrule
2  & 50.0   & 50.0   & 50.0   & 50.0   & 50.3 & 50.2 & 50.9 & 50.6 & 50.2 & 49.1 & 50.5 & 49.8 \\
4  & 48.9 & 50.1 & 50.5 & 47.0   & 50.4 & 50.4 & 54.7 & 51.9 & 52.9 & \cellcolor{lightgray}69.4 & \cellcolor{gray!25}64.8 & \cellcolor{gray!25}63.6 \\
6  & 49.8 & 50.5 & 49.7 & 48.0   & 49.2 & 50.2 & \cellcolor{gray!15}59.0   & \cellcolor{gray!15}58.0 & \cellcolor{gray!15}56.6 & \cellcolor{gray!25}62.7 & \cellcolor{gray!25}61.8 & \cellcolor{gray!25}61.4 \\
8  & 49.6 & 50.0   & 50.7 & 53.6 & 51.7 & 51.8 & \cellcolor{lightgray}66.1 & \cellcolor{gray!25}64.8 & \cellcolor{gray!25}61.3 & \cellcolor{lightgray}69.2 & \cellcolor{lightgray}66.3 & \cellcolor{gray!25}64.8 \\
10 & 50.3 & 51.8 & 49.9 & \cellcolor{gray!25}62.5 & \cellcolor{gray!15}57.7 & 54.3 & \cellcolor{lightgray}66.3 & \cellcolor{lightgray}65.6 & \cellcolor{gray!25}64.6 & \cellcolor{lightgray}68.5 & \cellcolor{gray!25}64.6 & \cellcolor{gray!25}62.9 \\
12 & 49.9 & 50.5 & 50.0   & \cellcolor{gray!15}55.1 & 51.4 & 51.3 & \cellcolor{lightgray}69.3 & \cellcolor{lightgray}68.1 & \cellcolor{lightgray}67.7 & \cellcolor{gray}74.1 & \cellcolor{lightgray}68.3 & \cellcolor{lightgray}65.1 \\
\midrule
Avg. & 49.8 & 50.4 & 50.1 & 52.7 & 51.8 & 51.4 & \cellcolor{gray!25}61.1 & \cellcolor{gray!15}59.8 & \cellcolor{gray!15}58.9 & \cellcolor{lightgray}65.5 & \cellcolor{gray!25}62.7 & \cellcolor{gray!25}61.3 \\
\bottomrule
    \end{tabular}
    \caption{Accuracy for all models on GELP based on the three degrees of memory load.} \label{table_accuracy_by_memory_cost_appendix}
\end{table*}

\section{Model Accuracy by Memory Load}\label{accuracy_by_memory_cost_appendix}
Table~\ref{table_accuracy_by_memory_cost_appendix} presents full results on model accuracy by memory load.

\end{document}